Bhat Mohammad Idrees [*] and B. Sharada.

# Spectral Graph-based Features for Recognition of Handwritten Characters: A Case Study on Handwritten Devanagari Numerals

**Abstract:** Interpretation of different writing styles, unconstrained cursiveness and relationship between different primitive parts is an essential and challenging task for recognition of handwritten characters. As feature representation is inadequate, appropriate interpretation /description of handwritten characters seems to be a challenging task. Although existing research in handwritten characters is extensive, it still remains a challenge to get the effective representation of characters in feature space. In this paper, we made an attempt to circumvent these problems by proposing an approach that exploits the robust graph representation and spectral graph embedding concept to characterize and effectively represents handwritten characters, taking into account writing styles, cursiveness and relationships. For corroboration of the efficacy of the proposed method, extensive experiments were carried out on standard handwritten numeral CVPR Unit, ISI, Kolkata Dataset. The experimental results demonstrate promising findings, which can be used in future studies.

## 1 Introduction

Optical Character Recognition (OCR) is concerned with automatic recognition of scanned and digitized images of text by a computer. These scanned images of text undergo various manipulations and then encoded with character codes such as American Standard Code for Information Interchange (ASCII), Unicode etc. OCR system tries to bridge the communication gap between man and machine and aides in automation of office with saving of considerable amount of time and human effort. Despite decades of research on different issues related to OCR [1, 2], research on handwritten characters has been less than satisfactory. It is an essential and challenging task for the community of pattern recognition. It is primarily because of the absence of fixed structure, the presence of numerous character shapes, cursiveness, the difference in inter and intra writer styles. Potential practical applications of it included in the automatic reading of postal codes, bank cheques, employee id, data entry, zip codes etc. Thus, recognition of handwritten characters is still an open area of research. In general, problems are associated with all handwritten documents. In this paper, we have considered a case study of handwritten Devanagari numerals, because of its importance in the Indian context.

One important question is how to give adequate representation/description of the underlying object (handwritten character) such that any recognition algorithm can be applied. Representation of object is done through two ways, viz. statistical representation and structural representation. In a statistical representation, the character is represented as a feature vector comprising of '$n$' measurements or values and can be thought as a point in $n-$ dimensional vector space, i.e. $F = (f_1, f_2, \ldots, f_n) \in R^n$. However, it has two representational limitations viz. dimension is fixed a priori i.e. all vectors in a recognition system have to agree with same length irrespective of the varying size of the underlying objects and second, they are inadequate in representing binary relationships that exist in primitive parts of the underlying object. Despite these, they are extensively used due to their flexible and computationally efficient mathematical base. For example, sum, product, mean etc, which are basic artefacts for many pattern recognition algorithms, can easily be computed. On the other hand, structural representation is based on symbolic data structure viz. graphs. The aforementioned limitations of feature vectors can be circumvented by graph representation [3, 4]. However, little algebraic support (less mathematical flexibility) and computationally expensive nature of many algorithms are major drawbacks to it. Compared to feature representation method, graphs provide robust representation formalism for the description of two-dimensional nature of handwritten characters viz. style variance, shape transformations, cursiveness, and size variance [4]. In this work, in order to exploit advantages of both, we have given graph representation to handwritten numerals to capture different writing styles, cursiveness and size variability. Thereafter, graphs are transformed into vector space by the concept of Spectral Graph Theory (*SGT*) to characterize the numeral graphs. Rest of the paper is organized into five sections: - Section 2 gives brief literature on the handwritten Devanagari numeral recognition system. An overview of definitions/illustrations of the terminologies used with respect to graph and spec-tral graph theory is given in section 3. In section 4 details about the proposed system is given. The recognition experiment is described in section 5, starting with a description of the dataset and experimental setup, followed by experimental results and concluded by a comparison with related work. Finally, Future work and conclusion is drawn in section 6.

## 2 Related Works

Over the years, an enormous amount of research work has been carried out in an attempt to make OCR, a reality. Different studies have explored various techniques like template matching [5], multi-pass hybrid method [6], syntactic features [7], shadow based features [8, 9] , gradient features [10,11 ] and CNN based features [12], to name just a few. Robust and stable features that are discriminating in feature space are an indispensable component in any recognition system. Inevitable characteristic of such features is that they should withstand to different types of variations (style, size etc) and shape transformations viz., rotation, scale, translation and reflection. Selection and extraction of such features in handwritten characters in the Indian context have been attempted by a number of researchers.

In Ref. [13], moment features (left, right, upper, and lower profile curves), descriptive component features and density features are combined for neural network based architecture for recognition. The main aim of extracting these types of features is to capture different stylistic variations. In Ref. [14], after giving wavelet-based multi-resolution representation, a numeral is subjected to the multi-stage recognition process. In each stage, a distinct Multi-Layer-Perceptron (MLP) classifier is used which either performs recognition or rejection. Thereafter, recognition for rejected numeral is attempted at next higher level. A fuzzy model-based system is proposed in Ref. [15], numerals are represented in the form of the exponential membership function, which behaves as a fuzzy model. Later recognition is performed by modifying exponential membership functions fitted to the fuzzy sets. Fuzzy sets are extracted from features comprising of normalized distances using the Box approach. An attempt is made in Ref. [16], to extract moment invariant features based on correlation coefficient, perturbed moments, image partitions and principal component analysis (PCA). These features are then used with Gaussian distribution function (GDF) for recognition purpose. In Ref. [17], translation and scale invariance of numerals are achieved by exploiting geometric moments such as Zernike moments. Extensive experiments were carried out on a large dataset that revealed the robustness of the proposed model. After giving graph representation different graph matching techniques are utilised such as sub-graph isomorphism, maximum common sub-graph and graph edit distance for Holistic recognition of Devanagari word [18], Oriya digit [19], and Devanagari numerals [20], respectively. However, the robustness of the graph representation is overshadowed by time complexity in these approaches.

A novel scheme based on edge histogram features is proposed in Ref. [21], scanned numeral images are pre-processed with splines together with PCA in order to improve the recognition performance. A local-based approach is proposed in Ref. [22], which exploits 16-segment display concept, extracted from half-toned binary images of numerals. A novel approach for recognizing handwritten numerals of five Indian sub-continent scripts has been proposed in Ref. [23]. Handwritten numerals are characterized by a combination of features such as Principal component analysis (PCA)/Modular PCA (MPCA) and Quadtree-based hierarchically derived longest run (QTLR).The efficacy of the proposed approach is validated by conducting extensive experiments on various datasets and the results demonstrate significant development in recognition performance. A global-based approach proposed in Ref. [24], in which features are extracted from end-points of numeral images. Thereafter, Recognition is carried with the neuromagnetic model. The feature level fusion based approach is attempted in Ref. [25], in which global and local features are combined together for Artificial Neural network based recognition. Several techniques gained importance due to their performance such as chain code features [26], Feature sub-selection [24], Zernike moments [27], and Structural features [28]. For a comprehensive survey, we refer readers to [29-31].

From the literature survey, we observe many researchers have addressed the problem of handwritten Devanagari numeral recognition by addressing separate objectives (shape transformations, style variations etc.). However, no attempts were made to address the problem as a whole. As numerals written by people are with different writing styles, even variation of style exist within-writer also; handwritten numeral recognition seems to be difficult and challenging. Thus, there is a scope for various attempts in this direction. Also, the reported works clearly indicate that the attempts have been made only by giving feature representation. However, as stated earlier, feature representation implicates two limitations namely size constraint and inability to represent binary relationships. These two limitations are severe in representing inherent two-dimensional nature of handwriting. With this observation, if these two limitations can be removed from recognition systems, greater and reliable recognition accuracies can be achieved. Hence, there is a scope to devise a model to circumvent stated limitations by providing robust alternative representation. From such a representation, besides representing object properties, we expect that inherent two-dimensional information is adequately modelled and binary relationships are preserved.

Graph representation models dependencies and relations among different primitive parts (by edges). Moreover, describing object properties. Furthermore, flexible in representing different individual object size in an application and invariant to shape transformations (scale, rotation, translation, reflection and mirror image) as well [32]. These characteristics of graphs are extremely beneficial to cope with different writing styles and cursiveness. Also, from the survey, with different applications such as image classification [33], image segmentation [34], synthetic graph classification [35], and many more, we observe Spectral Graph Theory (SGT) is more effective to characterize the graphs under consideration. SGT is a branch of mathematics that is primarily concerned with describing the structural properties of graphs by extracting eigenvalues of different graph associated matrices. The eigenvalues form the spectrum of the graph and exhibit interesting properties which can be exploited for recognition purposes. To enhance the recognition performance classifier fusion at decision level is also utilized. CVPR, Unit, ISI Kolkata dataset is employed as a dataset due to its popularity, availability and its complexity. Recognition results are lesser than the best result claimed by [36]. However, the main aim was not to outperform it but to circumvent stated limitations by giving graph representation and observe the results.

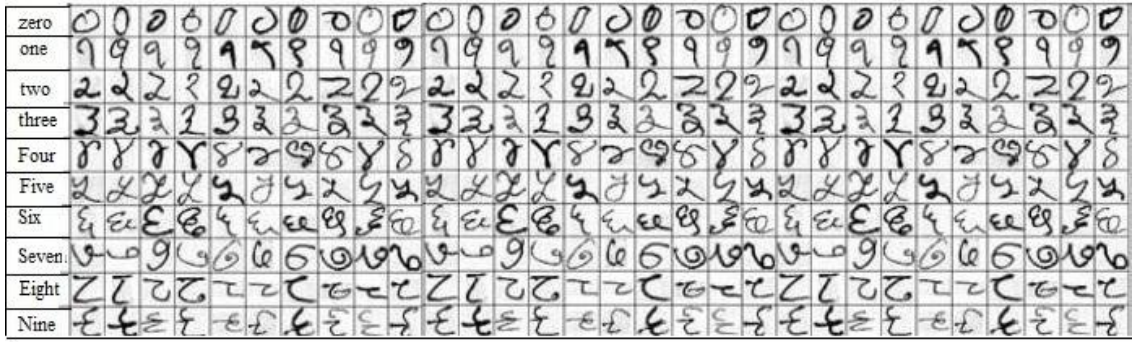

**Fig. 1:** Illustration of numeral images with several intra-class variations with respect to size and style.

## 3 Required Graph Terminologies

Brief and concise illustrations are given for various terminologies used in this study vis-a-vis graph theory and spectral graph theory (SGT). However, for comprehensive reading, we refer readers to [37-40].

**Definition 1. (Graph).** A Graph is a four-tuple $G = (V, E, \mu, \nu)$ where,

$V$                       set of vertices (or nodes); cardinality of it, is the order of the graph.

$E \subseteq V \times V$         set of Edges; cardinality of it is the size of the graph.

$\mu : V \to l_v$           *associating labels,* $l_v$, *to each vertex in* $V$.

$\nu : E \to l_e$           *associating labels,* $l_e$, *to each edge in* $E$.

A *directed graph* or *digraph* $G$ in which all edges $e$ in $E$ are directed from one vertex to another i.e., vertices are ordered pairs in $V$. An *undirected graph* $G$ is a graph in which all edges $e$ in $E$ are bidirectional i.e., vertices are unordered pairs in $V$. A weighted Graph $G$ is a graph in which each edge $e$ in $E$ is assigned a numerical weight by some *weighting function* $w(e_i)$. Mainly non-negative numeric values are used (called as the cost of the edges). One such weighting function $w(e_i)$ is the length of the edge $e$ in $E$. The degree of a vertex $v$ denoted by $d(v)$ in $G$ is the total number of vertices that are adjacent to it. There are different matrices associated with graphs which are important such as Adjacency matrix, Laplacian matrix. In a Graph $G$ with $|V|$ vertices, an adjacency matrix ($A(G)$) is a $|V| \times |V|$ matrix. Each $a_{ij}$ in $A(G)$ is 1 if the vertices $\{v_i, v_j\}$ in $V$ are adjacent, otherwise 0. *Laplacian matrix ( $L(G)$ )* of Graph $G$ is defined as $L(G) = D(G) - A(G)$, where $D(G)$ and $A(G)$ are the degree and Adjacency matrix of Graph $G$. Each $l_{ij}$ in $L(G)$ is $\deg(v_i)$ if $\{v_i = v_j\} \forall i, j$, -1 if edges $e$ in $E$ are adjacent ($\forall i \neq j$) and 0 otherwise. *Weighted Adjacency Matrix, $WA(G)$* is constructed by removing all entries where $\{v_i, v_j\} = 1$ in $A(G)$ with respective weights assigned by a weighting function $w(\{v_i, v_j\})$. Weighted Laplacian matrix $WL(G) = D(G) - WA(G)$ where $D(G)$ is a degree matrix each $l_{ij}$ in $WL(G)$ is $\deg(v_i)$ if $\{v_i = v_j\} \forall i, j$, negative times weight assigned by a weighting function to edges $e$ in $E$ which are adjacent ($\forall i \neq j$) and 0 otherwise. *Distance matrix, $Dist(G)$* of vertices in a graph $G$ is $|V| \times |V|$ matrix contains pairwise distances (provided by a weighting function, $w(e_i)$) between each $v$ in $V$ i.e., distances are included even for non-adjacent nodes $v$ in $V$. Despite robust structural representational formalism of objects, as stated earlier, graph-based methods in pattern recognition (like *graph matching*) have major limitations. These limitations include computationally expensive nature of algorithms and presence of little algebraic properties (basic operations required in many pattern recognition algorithms such as sum, mean, and product etc. are not defined in a standard way). In order to, overcome these limitations graphs are transformed into low dimensional vector space such a technique is called *Graph embedding* $\varphi: G \rightarrow R^n$. One such technique is *Spectral Graph Embedding* (SGE), in which graphs are transformed into vector space by the *Spectrum of the graph*. The spectrum of Graph $G$ (where $G$ can be represented by any graph associated matrix $M$, in this study $(WA(G), WL(G), and\ Dist(G))$ is the set of *eigenvalues*, together with their algebraic multiplicities (number of times they occur). Representation of any graph associated matrix in terms of its *eigenvalues* and *eigenvectors* is called its *eigendecomposition/spectral decomposition*. For better illustration, let $G(5,7)$ be the graph in which each edge $e$ is *weighted (labelled)* arbitrarily, and then the desired matrices can be extracted as shown in Fig.2. It should be noted here that there is a subtle difference between label and weight of the graph, in this study label and weight refer the same and are used interchangeably.

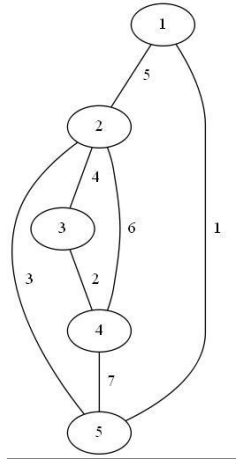

$$WA(G) = \begin{bmatrix} 0 & 5 & 0 & 0 & 3 \\ 5 & 0 & 4 & 6 & 3 \\ 0 & 4 & 0 & 2 & 0 \\ 0 & 6 & 2 & 0 & 7 \\ 1 & 3 & 0 & 7 & 0 \end{bmatrix} \quad D(G) = \begin{bmatrix} 2 & 0 & 0 & 0 & 0 \\ 0 & 3 & 0 & 0 & 0 \\ 0 & 0 & 2 & 0 & 0 \\ 0 & 0 & 0 & 3 & 0 \\ 0 & 0 & 0 & 0 & 3 \end{bmatrix} \quad WL(G) = \begin{bmatrix} 2 & -5 & 0 & 0 & -1 \\ -5 & 3 & -4 & -6 & -3 \\ 0 & -4 & 2 & -2 & 0 \\ 0 & -6 & -2 & 3 & -7 \\ -1 & -3 & 0 & -7 & 3 \end{bmatrix}$$

**Fig. 2:** Weighted Graph $G$ (5, 7) (order $|V|=5$ and size $|E|=7$, labelled arbitrarily) and its associated weighted Adjacency $WA(G)$, Degree Matrix $D(G)$, and weighted Laplacian Matrix $WL(G)$ respectively. $(WL(G) = D(G) - WA(G))$.

## 4 Proposed Model

Various steps involved in the proposed handwritten Devanagari numeral recognition model are shown in Fig.3. These steps are explained in following subsections**:-**

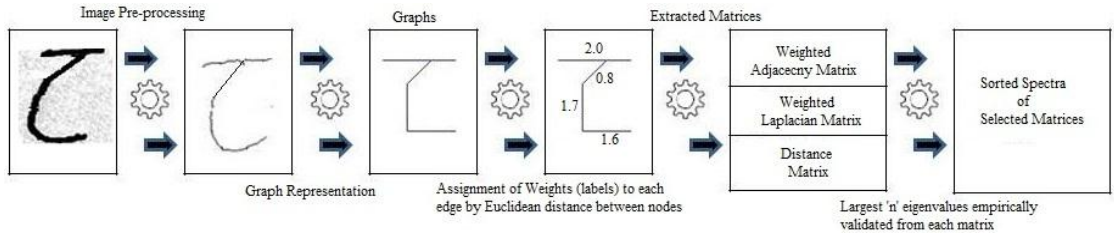

**Fig. 3:** Process of extraction of sorted spectra.

## 4.1 Image Pre-processing

Image pre-processing deals with reducing variations on scanned images of handwritten numerals caused by noise. In this study, scanned numeral images are first filtered by Difference of Gaussian (*DoG*)-filtering, second normalization is applied to handle variability in size, and later numeral images are binarized. Finally, numeral images are skeletonised by a $3 \times 3$ thinning operator [41].

## 4.2 Graph Representation

There exist various graph representations [32], however, we selected interest point graph representation as it preserves inherent structural characteristics of numeral images. It identifies the points in an image where the signal information is rich such as junction points, start and end points, corner points of circular primitive parts of numerals. Fig.4 shows some extracted sample numeral graphs and interest points in each numeral graph.

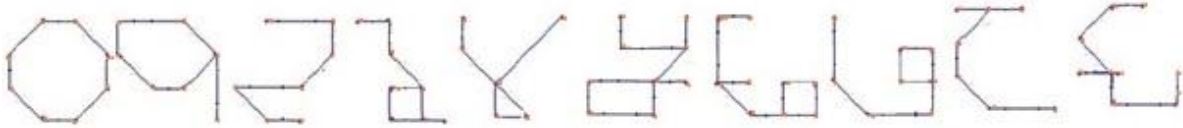

**Fig. 4:** Snapshot of underlying graphs obtained from handwritten Devanagari numerals with interest points (0-9)

## 4.3 Feature Extraction

Weighted graphs include more discriminating information than un-weighted such as stretching of the graph [32]. In order to give weights to numeral graphs, edges are labelled with most well-known and intuitive weighting function $w: E(G) \rightarrow \mathbb{Z}^+$ which assigns Euclidean distance to each edge in $G$. Euclidean distance is computed from respective $2D$ coordinates of nodes incident with each edge $e$ in $E$ (shown in Fig.5a). The motivation behind using such a weighting function is twofold; first, it is computationally simple and, secondly, the distance between any two objects (in this study, nodes) remains unaffected with the inclusion of more objects (nodes) in the analysis [42]. However, there is an arsenal of weighting functions described in the literature [43] one can use any one of them. As stated earlier, spectral graph embedding is described in terms of matrices associated with graphs. Selection and extraction of matrices which preserve underlining structure or topology of the numeral graphs are indispensable. In consideration of this fact, we selected weighted Adjacency matrix ($WA(G)$), weighted Laplacian matrix ($WL(G)$) and Distance Matrix ($Dist(G)$). These matrices exhibit different topological information (global or local) of graphs which can be crucial for the characterisation of numeral graphs. Adjacency matrix consists of a length of edges and it is unique for each graph (up to permutation rows and columns) that leads to isomorphism, invariance of graphs. A total number of connected components and spanning trees for a given graph is given by Laplacian matrix. A number of spanning trees $t(G)$, in a connected graph, is a well-known invariant and leads to many more discriminating properties of the graph. Distance matrix gives the mutual pair-wise distance between each node; the matrix thus formed is different for graphs having equal order [44-47]. Matrix decomposition follows subsequent representation of these matrices in-terms of eigenvalues (with their multiplicities) called spectral decomposition or eigendecomposition of graphs. Let $M$ be some matrix representation of graph $G$ ($WA(G)$, $WL(G)$ and $Dist(G)$) then the spectral decomposition (or eigendecomposition) is $M = \Phi \Lambda \Phi^T$ where $\Lambda = diag(\lambda_1, \lambda_2, \lambda_3, \ldots, \lambda_{|V|})$ is the ordered eigenvalues of a diagonal matrix and $\Phi = (\Phi_1, \Phi_2, \Phi_3, \ldots, \Phi_{|V|})$ is

the ordered eigenvectors as columns in a matrix $M$. Then the spectrum (eigendecomposition) of $M$ is the set of eigenvalues $\{\lambda_1, \lambda_2, \lambda_3, \ldots, \lambda_{|V|}\}$. For the eigenvalues $\{\lambda_1, \lambda_2, \lambda_3, \ldots, \lambda_{|V|}\}$ and corresponding eigenvectors $(\Phi_1, \Phi_2, \Phi_3, \ldots, \Phi_{|V|})$ equation (1) holds. The advantage of using spectrum in characterising a graph is that eigendecomposition of various matrices associated with graphs can be quickly computed (computation of a spectrum from a matrix requires $O(n^3)$ operations, where '$n$' is the order of the graph). Furthermore, Spectral parameters of graph illustrate/specify various discriminating properties, that otherwise are exponentially computed (chromatic number, sub-graph isomorphism, perturbation of graph, number of paths of length '$K$' between two nodes, number of connected components in a graph etc.). Thus exploiting spectrum for the graph characterisation is clearly beneficial.

$$M\Phi = \lambda\Phi \qquad (I)$$

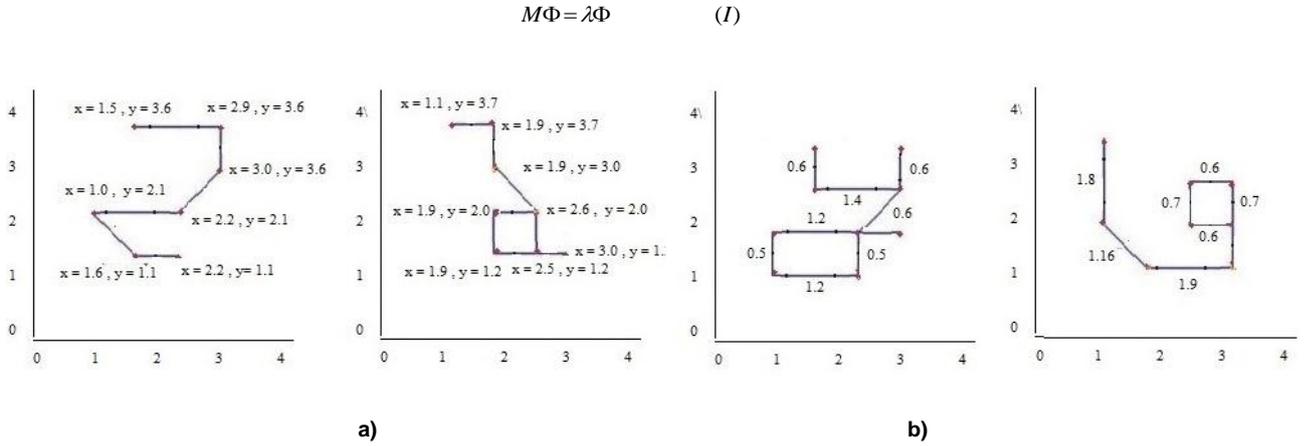

a)          b)

**Fig. 5:** Illustration of assigning weights to numeral graphs a) each node labelled with 2D coordinates b) each edge in numeral graph labelled (weighted) with the Euclidean distance between two adjacent nodes.

For an illustration of eigendecomposition, let $WAG = M$ be the matrix representation of a graph $G$ described in section 3.

Equation (1) can also be written as

$$M\Phi - \lambda I\Phi = 0 \;\Rightarrow\; (M - \lambda I)\Phi = 0 \;\Rightarrow\; \det(M - \lambda I) = 0 \qquad (II)$$

Where '$I$' is the identity matrix, $\Phi$ is a special vector (eigenvector) that is in the same direction as $M\Phi$. After multiplying $\Phi$ with $M$, the vector $M\Phi$ is a number $\lambda$ time the actual $\Phi$, called as an eigenvalue of $M$. That means, upon linear transformation $M$ on $\Phi$, $\lambda$ is an amount of how much vector $\Phi$ is elongated or shrunk, reversed or unchanged, that is described by, an eigenvalue.

Eigendecomposition of weighted Adjacency Matrix ($WAG$) can be carried out as follows:-

$$WAG = \begin{bmatrix} 0 & 5 & 0 & 0 & 1 \\ 5 & 0 & 4 & 6 & 3 \\ 0 & 4 & 0 & 2 & 0 \\ 0 & 6 & 2 & 0 & 7 \\ 1 & 3 & 0 & 7 & 0 \end{bmatrix} \quad \text{after applying } (II) \quad \begin{bmatrix} -\lambda & 5 & 0 & 0 & 1 \\ 5 & -\lambda & 4 & 6 & 3 \\ 0 & 4 & -\lambda & 2 & 0 \\ 0 & 6 & 2 & -\lambda & 7 \\ 1 & 3 & 0 & 7 & -\lambda \end{bmatrix}$$

Then solving for the equation, $-\lambda + 140\lambda^3 + 378\lambda^2 - 1445\lambda + 344$, we arrive at ordered (dominant) eigenvalues:-

$$\Lambda = (12.6880, 1.9669, 0.2570, -6.0595, -8.8.523)$$

Similarly, eigendecomposition is carried out for Weighted Laplacian Matrix $WL(G)$ and Distance Matrix $Dist(G)$. Thereafter, we arrive at feature matrices consisting of ordered (dominant) eigenvalues (spectrum) of $WAG$, $WL(G)$, $Dist(G)$ respectively. Further, these features (spectrum) are first inspected individually for characterisation potential and later they are fused together at decision level (or classifier level fusion) to characterise the numeral graphs.

### 4.4 Adequacy of the Features

Spectrum inherits different properties (global and local) from their respective graph associated matrices which make them ideal candidates for recognition purposes, a thorough study can be found in [48-51]. However, few important properties which are concerned with this study are described as follows:-
- Spectrum is real if the associated graph matrix is real and symmetric. Hence, the spectral decomposition map graphs in a coordinate system thereafter, any clustering or classification procedures can be used.
- Spectrum is invariant with respect to labelling of the graph (Isomorphic graphs ) if sorted either in ascending or descending order because swapping of two columns have no effect on values. Therefore, different orders of the graphs have no influence.
- Since each eigenvalue contains information about all nodes in a graph so it is possible to use an only certain subset of them. Therefore, it is not mandatory to use all eigenvalues. Imbalanced (short) spectra's can be balanced with padding zero values.
- For disconnected graph $G$ spectrum is the union of the spectra's of different components in $G$.

## 5 Experimentation

### 5.1 Dataset Description and Experimental Setup

For experimentation, we have used isolated handwritten Devanagari numeral dataset from Computer Vision Pattern Recognition, Unit of Indian Statistical Institute Kolkata (CVPR, Unit, ISI Kolkata). It consists of 22,556 samples written by 1049 persons. 368 mail pieces, 274 job application forms, and specially designed forms were used. In a dataset, numerals are with different writing styles, size and stroke widths. Dataset also comprises of certain samples that cannot be recognized by humans also. We divided entire dataset of labelled numeral images into three disjoint sets viz. Training, validation and test set respectively. The validation set is used to tune/optimise the meta-parameters of the classifier and proposed method. However, original dataset is divided into training and testing ratios, but the authors of the dataset have stated in [52], that depending upon the requirement, the dataset can be partitioned into training, validation and test sets, respectively. Hence, we divided the dataset into two standard ratios of 60:20:20 and 50:25:25 [53] of training, validation and test set respectively. Fig.1 shows some numeral samples of the dataset. The complete description of the dataset can be found in [54].

Due to its robustness, which is validated from numerous fields of pattern recognition, we have employed multi-class Support Vector Machines in association with a kernel called Gaussian Kernel (also called Radial basis function, $RBF-$ kernel) [56,57]. There are two possible ways of classification in multi-class $SVM$ : One-vs.-One classifier ( $IV1$ ) and One-vs.-All classifier ( $IVA$ ). We have utilized One-vs.-One method, as it is insensitive towards imbalanced dataset. In this method, training is done with all pairs of two-class $SVMs$ (e.g., for 3 class problem, $1-3, 2-3, 1-2$ ) also called pair-wise decomposition. All possible pair-wise classifiers ( $n(n-1)/2$ ) are evaluated and decision for unseen observation is made by majority vote. During training $RBF$ based $SVM$ have to optimise, two meta-parameters (namely $C$ and $\Upsilon$ , represents classification cost and non-linear function respectively), empirically on the dataset. To arrive at optimised parameters, values for $C$ and $\Upsilon$ are varied from $0.001-10,000$ on a logarithmic scale ( $base-2$ ) ($i.e.$, $0.001, 0.01.........)$ . Each $SVM$ is trained for every possible pair ( $C$ , $\Upsilon$ ) on the training set and the recognition accuracy is tested on the validation set. Values leading to the best recognition accuracy are then used with Independent test set (Table I).

Each spectrum (spectra of $WA(G)$ , $WL(G)$ , and $Dist(G)$ ) is investigated individually for recognition potential. From now on, we refer spectra of $WA(G)$ , $WL(G)$ ¸ and $Dist(G)$ as (Feature Type) $FT_1, FT_2, and\ FT_3$ respectively. The individual recognition results from each feature type are then compared. In order to improve the accuracy of individual classifiers, Multi-classifier system (MCS) [57] or classifier fusion is employed. Classifier fusion combines their results by using various combining strategies; however we used Bayesian fusion (described in subsection 5.2). It is worth underlining that in MCS, individual classifiers should be accurate and diverse [57]. As stated earlier, the accuracy of $SVMs$ is experimentally validated in a number of practical recognition problems, diversity means each classifier should make different errors or their decision boundaries should be different. In this study, diversity is achieved by using different feature types (as discussed in sub-section 4.3) of the numeral graphs.

## 5.2 Fusion Technique

We used Bayesian combination rule (also known as Bayesian Belief Integration) as a combined technique. It is based on the concept of conditional probability. To compute the conditional probabilities of each classifier for all classes, confusion matrix has to be calculated first. Let $C^l$ be the confusion matrix for each classifier $e_l$ with $l=1,...,L$, where $L$ is the total number of classifiers used (in this study $L=3$).

$$C^l = \begin{bmatrix} C_{11} & C_{12} & \cdots & C_{1N} \\ C_{21} & C_{22} & \cdots & C_{2N} \\ C_{31} & C_{31} & \cdots & C_{3N} \\ \vdots & \vdots & \ddots & \vdots \\ C_{N1} & C_{N2} & & C_{NN} \end{bmatrix} \qquad (2)$$

Where, $i, j = 1,...,N$, $N$ is the number of classes, $C_{i,j}$ in $C^l$ is the total number of samples in which classifier $e_l$ predicted class label $j$ whereas actual label was $i$. By using information present in confusion matrix the probability, that the test sample '$x$' corresponds to class '$i$' if the classifier $e_l$ predicts class $j$ can be calculated as follows:-

$$P_{ij} = P(x \in i \mid e_l(x) = j) = \frac{C_{i,j}^l}{\sum_{i=1}^{N} C_{i,j}^l} \qquad (3)$$

Probability matrix $P^l$ for each classifier $e_l$ is:-

$$P^l = \begin{bmatrix} P_{11} & P_{12} & \cdots & P_{1N} \\ P_{21} & P_{22} & \cdots & P_{2N} \\ P_{31} & P_{32} & \cdots & P_{3N} \\ \vdots & \vdots & \ddots & \vdots \\ P_{N1} & P_{N2} & \cdots & P_{NN} \end{bmatrix} \qquad (4)$$

Based on $P^l$ for each classifier a combined estimate value, $b(i)$ for each class '$i$', is calculated for each sample '$x$' in test set.

$$b(i) = \frac{\prod_{l=1}^{L} P_{i,jl}}{\sum_{i=1}^{N} \prod_{l=1}^{L} P_{i,jl}} \qquad (5)$$

For a test sample, 'x' classifier $e_l$ predicts class label $j_l$. To make a decision for one of the class maximum values in $b(i)$ is used.

## 5.3 Experimental Results

Several experiments were carried out for all three feature types ($FT_1, FT_2,$ and $FT_3$) and subsequently repeated for 50 random trials of training, validation and testing in the ratios of $60:20:20$ and $50:25:25$ respectively. In each trial, the performance of the proposed method is assessed by the recognition rate in terms of $F-$measure and the average $F-$measure is computed from all trials. Table 1 gives the class wise performance in-terms of $F-$measure (for both the ratios belonging to all the feature types) and also presents validated meta-values for $RBF-kernel$. Fig.6 shows confusion matrices obtained for optimised parameters of the classifier (for each feature type ($FT_1, FT_2,$ and $FT_3$) respectively).

The performance of any recognition method is assessed in terms of precision, recall, and $F-measure$ described as follows:

$$\text{Precision} = \frac{CP}{CP+FP} \quad (6)$$

$$\text{Recall} = \frac{CP}{CP+FN} \quad (7)$$

$$\text{F-measure} = \frac{(2*\text{Precision}*\text{Recall})}{(\text{Precision}+\text{Recall})} \quad (8)$$

Measures Precision, Recall and $F-measure$ are based on correct positive, false negative, false positive, and correct negative for overall samples of the test set.

Table 2 presents the average $F-measure$ computed from all trails. Individually these feature types ($FT_1, FT_2,$ and $FT_3$) generate $75-85\%$ average recognition rate. Since $FT_3$ comprises of all the pair-wise distances, the shape of the numeral graph is not preserved. Numeral graphs with an equal number of vertices $|V|$, are only distinct in pair-wise distances of the vertices but equal in a number of non-zero entries. Perhaps, this could be the reason for its ($FT_3$) lowest recognition result ($75-76\%$). $FT_1$ & $FT_2$ preserve the exact shape of the numeral graphs such as the presence of edges and also their weights hence they generate over $80\%$ average recognition rates. Since each graph associated matrices contain non-overlapping information therefore by combining the classifiers at decision level greater recognition rates can be achieved. With classifier fusion at decision level, we achieved maximum average recognition rate (fusion is carried out individually for each trial and then average recognition accuracy is recorded) $93.73\%$ shown in Table 3. Therefore, by decision fusion at classifier level recognition rate is increased ($FT_1, FT_2,$ and $FT_3$) by $7.9\%$. The numerals which have same underlying graph structure (more or less) build the misclassified pairs such as Devanagari zero and Devanagari one (as can be observed from Table 1 and confusion matrices). Furthermore, invariance property of spectrum also adds to the confusion. It can be understood by observing the shape of the Devanagari numeral three and Devanagari numeral six (as shown in Fig.2, just mirror images of each other). Since we sorted spectrum, therefore, their spectra are more or less equal. In consideration of these facts, recognition performance is encouraging.

It should be noted that each spectrum was sorted in descending order. In order to choose $'n'$ largest eigenvalues for each feature type $FT_1$, $FT_2$ and $FT_3$, we have conducted experiments for various values $'n'$ on the validation set. We observe that only small value of $_{has}$ significant development ($n=3$). But when we increase the value of $'n'$ we have not observed much significant development in recognition performance. Thus, in experimentation, we have considered the value of $'n'$ equal to $3$ for every feature type ($FT_1, FT_2,$ and $FT_3$) respectively. The results obtained after fusion with varying $'n'$ are shown in Table 4.

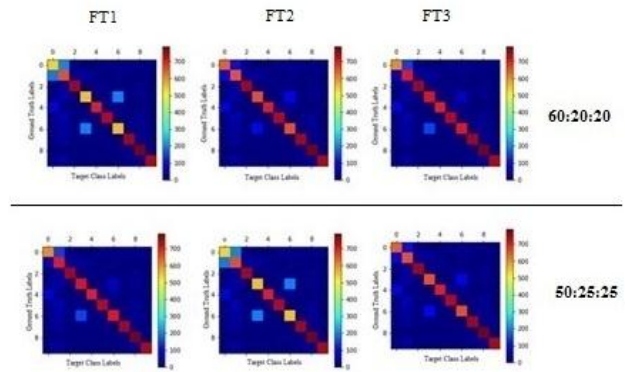

**Fig. 6:** Confusion matrices for each feature type ($FT_1, FT_2,$ and $FT_3$) for both divisions respectively

**Table. 1:** Class-wise performance of all feature types ($FT_1$ $C=0.125$ $\Upsilon=0.001$, $FT_2$ $C=0.031$ $\Upsilon=0.0004$, and $FT_3$ $C=0.001$ $\Upsilon=0.004$)

| Class Index | Training: Validation: Testing | | | | | | Class Index | Training: Validation: Testing | | | | | |
|---|---|---|---|---|---|---|---|---|---|---|---|---|---|
| | 60:20:20 | | | 50:25:25 | | | | 60:20:20 | | | 50:25:25 | | |
| | $FT_1$ | $FT_2$ | $FT_3$ | $FT_1$ | $FT_2$ | $FT_3$ | | $FT_1$ | $FT_2$ | $FT_3$ | $FT_1$ | $FT_2$ | $FT_3$ |
| 1 | 0.90 | 0.93 | 0.79 | 0.89 | 0.92 | 0.96 | 6 | 0.75 | 0.74 | 0.75 | 0.74 | 0.72 | 0.74 |
| 2 | 0.92 | 0.94 | 0.77 | 0.87 | 0.93 | 0.93 | 7 | 0.68 | 0.81 | 0.80 | 0.67 | 0.80 | 0.79 |
| 3 | 0.78 | 0.72 | 0.73 | 0.72 | 0.71 | 0.72 | 8 | 0.88 | 0.85 | 0.94 | 0.87 | 0.84 | 0.93 |
| 4 | 0.69 | 0.85 | 0.93 | 0.68 | 0.84 | 0.92 | 9 | 0.65 | 0.77 | 0.69 | 0.64 | 0.76 | 0.68 |
| 5 | 0.81 | 0.67 | 0.96 | 0.80 | 0.66 | 0.95 | 10 | 0.61 | 0.62 | 0.88 | 0.60 | 0.61 | 0.87 |

**Note**: Where $FT_1$ = feature type one or sorted spectrum of the weighted adjacency matrix, $FT_2$ = feature type two or sorted spectrum of weighted Laplacian matrix and $FT_3$ = feature type three or sorted spectrum of distance matrix respectively. Values of $C$ and $\Upsilon$ are the validated meta-parameters for $RBF-$ kernel $SVM$ for each feature type $FT_1, FT_2,$ and $FT_3$ respectively.

**Table. 2**: Overall Average Recognition Performance (in-terms of $F-$ measure) for both ratios.

| Dataset | Feature Type | Ratio's of Training, Validation and Testing | Overall Recognition Rate |
|---|---|---|---|
| CVPR-ISI Kolkata | $FT_1$ | 60:20:20 | 85.83±1.05 |
| | | 50:25:25 | 84.63±1.16 |
| | $FT_2$ | 60:20:20 | 83.93±0.98 |
| | | 50:25:25 | 82.73±0.86 |
| | $FT_3$ | 60:20:20 | 76.73±0.96 |
| | | 50:25:25 | 75.83±0.99 |

**Table. 3:** Average Recognition rate

| Dataset | Ratio's of Training, Validation and Testing | Average Recognition Rate in-terms of $F-$ measure |
|---|---|---|
| **CVPR-ISI Kolkata** | 60:20:20 | 93.83±1.12 |
| | 50:25:25 | 92.73±0.97 |

**Table. 4:** Empirical Evaluation of '$n$' Largest Eigen Values

| Ratio's of Training, Validation and Testing | Largest Eigen Values | Recognition Accuracy in-terms of $F-$ measure |
|---|---|---|
| 60:20:20 | 1 | 90.65±0.98 |
| | 2 | 91.75±0.95 |
| | 3 | 93.83±1.12 |
| | 4 | 89.85±0.93 |
| | 5 | 88.95±0.92 |
| 50:25:25 | 1 | 89.75±0.92 |
| | 2 | 90.85±0.96 |
| | 3 | 92.73±0.97 |
| | 4 | 86.75±0.91 |
| | 5 | 85.65±0.94 |

### 5.4 Comparative Study

We have compared our model with the paper, where graph representation is utilized on the same dataset. From the literature, we observe authors in [58] achieved recognition accuracy 95.85% (in terms of $F-$measure) by using Graph representation and Lipchitz embedding. Lipchitz embedding is based on transforming a graph into $'n'$ distances to already set aside $'n'$ m-dimensional reference set of graphs as shown in Fig.7. Each $'d_i'$ in Feature vector $F = (d_1, d_2, ..., d_n)$ is obtained by taking the minimum distance between input graph $'g'$ and graphs present in each reference set i.e., $d_i = \min(R_1, R_2, \ldots, R_m)$ where, $R_1, R_2, \ldots, R_m$ are the individual graphs belonging to each reference set.

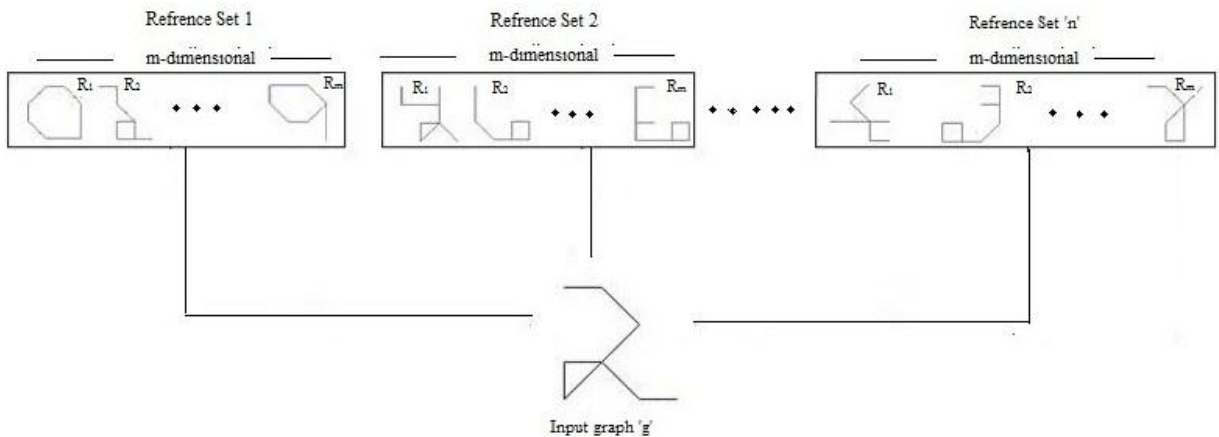

**Fig. 7:** Illustration of the compared model.

Consequently, a graph $'g'$ is converted to the $n-$dimensional vector space $R^n$ by computing the Graph Edit Distance (GED) of $'g'$ to all of the $'n'$ reference sets (each $m-$dimensional). However, transforming numeral graphs into vector spaces by computation of dissimilarities from $'n'$ $m-$dimensional selected reference set (carefully selected set of graphs) is time-consuming. Input graph $'g'$ is matched with every single graph in Reference set that requires time complexity in cubic order with respect to the order of the graph (thus inappropriate for graphs having large orders). Furthermore, GED depends on optimization of various factors viz. Insertion, deletion and substitution cost of nodes and edges. Recognition performance is greatly influenced by number and dimension of Reference sets. Moreover, type of the graphs selected from the dataset for each Reference set also has a great impact on the recognition performance. Our model is superior to discussed model despite its superior performance since we transform the numeral graph into vector space by eigendecomposition (or spectrum as a feature vector) to avoid computationally expensive graph matching.

Furthermore, most misclassification occurs in our model due to invariance property of the spectrum. Thus the efficacy of the proposed method can easily be justified. Since our model gives graph representation it is not directly comparable with conventional feature representation models.

## 6 Conclusion and Future Work

In this study, we presented a method that exploits robust graph representation and spectral graph embedding for recognition of style variant, cursive handwritten characters by taking a case study of Devanagari numerals. Largest $'n'$ eigenvalues (spectrum) are extracted from selected (application dependent) weighted numeral graph associated matrices. We empirically validated highest performing $'n'$ from each spectrum. Recognition performance from individual spectrum ranges from $75-85\%$ (in terms of average F-measure). In order to augment recognition accuracy classifier fusion at the decision level is also studied. That increases recognition accuracy significantly as shown in Table 4. Performance of the method is corroborated by conducting extensive experiments on standard CVPR-ISI Kolkata dataset. After observing, the results from different experiments, we conclude that the proposed method is effective in repre-senting complex relationships between different primitives, different intra-class size, style, image transformations (translation, scale, rotation, reflection and mirror image) and cursiveness for recognition of handwritten Devanagari numerals. However, the method may not withstand with handwritten characters/numeral if they have same (more or less) underlying graph representation. Furthermore, invariance property of the spectrum also adds to the confusion. Hence, due to these reasons, most misclassification occurs.

There are various issues that need further investigation. For example, there seems to be room for employing spectra of the further graph associated matrices at decision level fusion. Furthermore, experiments/observations in this study have been based on Support Vector Machines. It would be interesting to repeat experiments/observations with different classifiers. Moreover, utilizing probabilistic outputs (Fuzzy) in One-vs.-one and one-vs.-all Multi-class classification seems to be an interesting topic for further research. Finally, in this study, we have utilised Euclidean distance for labelling graphs. It would be interesting to observe the influence of distance on eigendecomposition of numeral graphs.

**Acknowledgement.** We would like to thank Prof. Ujjwal Bhattacharya and Prof. B.B. Chaudhuri of Computer Vision and Pattern Recognition Unit (CVPR-Unit) *of* Indian Statistical Institute (ISI) Kolkata for providing Handwritten Devanagari Numeral dataset.